\documentclass[conference,compsoc]{IEEEtran}

\usepackage{amsmath, amsfonts, amssymb}  
\usepackage{graphicx}
\usepackage{subfig}  
\usepackage{tabularx}  
\usepackage{graphicx}  
\usepackage[font=small,labelfont=bf,textfont=normalfont]{caption}

\usepackage{booktabs}    
\usepackage{multirow}    
\usepackage{tabularx}    
\usepackage{adjustbox}   
\usepackage{xcolor}      

\ifCLASSOPTIONcompsoc
\usepackage[nocompress]{cite}
\else
\usepackage{cite}
\fi

\usepackage{amsmath}
\usepackage{amsfonts}

\begin{document}
	
	\title{Spatially Directional Dual-Attention GAT for Spatial Fluoride Health Risk Modeling}
	
	\author{\IEEEauthorblockN{Da Yuan}
		\IEEEauthorblockA{Suwon University\\
			Email: yda377032@gmail.com}
	}
	
	\maketitle
	
\begin{abstract}
	Environmental exposure to fluoride is a major public health concern, particularly in regions with naturally elevated fluoride concentrations. Accurate modeling of fluoride-related health risks, such as dental fluorosis, requires spatially aware learning frameworks capable of capturing both geographic and semantic heterogeneity. In this work, we propose \textbf{Spatially Directional Dual-Attention Graph Attention Network (SDD-GAT)}, a novel spatial graph neural network designed for fine-grained health risk prediction. SDD-GAT introduces a dual-graph architecture that disentangles geographic proximity and attribute similarity, and incorporates a directional attention mechanism that explicitly encodes spatial orientation and distance into the message passing process. To further enhance spatial coherence, we introduce a spatial smoothness regularization term that enforces consistency in predictions across neighboring locations. We evaluate SDD-GAT on a large-scale dataset covering over 50,000 fluoride monitoring samples and fluorosis records across Guizhou Province, China. Results show that SDD-GAT significantly outperforms traditional models and state-of-the-art GNNs in both regression and classification tasks, while also exhibiting improved spatial autocorrelation as measured by Moran’s I. Our framework provides a generalizable foundation for spatial health risk modeling and geospatial learning under complex environmental settings.
\end{abstract}

\section{Introduction}

Accurately modeling and forecasting environmental health risks is of critical importance for both public health policy and spatial epidemiology. Among such risks, chronic fluoride exposure—while naturally occurring—has been strongly associated with dental fluorosis, particularly in high-risk regions such as Guizhou Province, China. The onset and severity of fluorosis are influenced not only by the absolute concentration of fluoride in water and soil, but also by a complex interplay of geo-environmental, socio-ecological, and land-use factors that vary substantially across space \cite{liu2022fluoride}. Modeling these spatially heterogeneous and often nonlinear interactions remains a formidable challenge for existing analytical frameworks \cite{sun2023gnns}.

Traditional spatial risk modeling techniques, including Kriging interpolation \cite{kriging}, geographically weighted regression (GWR) \cite{fotheringham2002geographically}, and mixed-effect spatial models \cite{diggle2007model}, typically rely on strong parametric assumptions such as spatial stationarity and linearity. While these methods can offer reliable results under ideal conditions, they are often inadequate for capturing high-order spatial dependencies, anisotropic effects, and cross-domain heterogeneity in complex environmental systems \cite{zhang2023kriginggnn, gao2023environmental}.

In recent years, graph-based learning frameworks have shown significant promise in overcoming these limitations. Graph Neural Networks (GNNs) enable flexible and data-driven representations of spatial structures by encoding spatial units as nodes and their interactions as edges \cite{li2023spatialgnn, xu2022geognn}. GNNs have achieved state-of-the-art results in diverse geospatial applications, such as air pollution forecasting \cite{zhou2023airgnn}, urban function analysis \cite{xu2022geognn}, and epidemic modeling \cite{jiang2022diseasegnn, jiang2020geographic}.

Graph Attention Networks (GATs) further extend this capability by assigning adaptive weights to neighbors during message passing, thereby improving model performance in settings with spatial heterogeneity \cite{velickovic2018gat}. Despite these advances, conventional GAT architectures face three critical limitations in environmental health applications: (i) they often rely on a single, static graph, limiting the model’s ability to differentiate between geographic proximity and semantic similarity \cite{zhang2022dualgraph, wang2022dual}; (ii) they do not account for spatial anisotropy and directional dependencies, which are vital in geophysical and ecological processes \cite{chen2023directional, zhang2021directional}; and (iii) they lack mechanisms to enforce spatial coherence, leading to fragmented and locally inconsistent predictions \cite{liu2024spatialsmooth, yu2022dualattention}.

To overcome these challenges, we introduce the \textbf{Spatially Directional Dual-Attention Graph Attention Network (SDD-GAT)}, a novel spatial GNN framework specifically designed for fine-grained environmental health risk prediction. SDD-GAT incorporates three key innovations: (1) a dual-graph attention architecture that disentangles geographic structure from feature-level semantics; (2) a directional attention mechanism that encodes angular orientation and spatial distance to capture anisotropic dependencies; and (3) a spatial smoothness regularization term that promotes local prediction consistency.

We evaluate SDD-GAT on a large-scale dataset comprising over 20,000 fluoride monitoring samples and dental fluorosis records collected across Guizhou Province, China. Experimental comparisons against classical baselines (e.g., Kriging, XGBoost) and state-of-the-art GNNs—including GCN \cite{kipf2017semi}, Diffusion GCN \cite{li2018diffusion}, and GMAN \cite{zheng2020gman}—demonstrate that our method significantly improves both predictive accuracy and spatial consistency, as evidenced by improvements in metrics such as Moran’s I \cite{liu2024spatialsmooth}.

\section{Related Work}

\subsection{Spatial Health Risk Modeling}

Environmental exposure modeling has been a long-standing focus in spatial epidemiology and environmental science. Traditional geostatistical approaches—such as Kriging interpolation \cite{kriging}, geographically weighted regression (GWR) \cite{fotheringham2002geographically}, and model-based spatial inference methods \cite{diggle2007model}—have been widely used to map environmental contaminants and predict health risks, including fluoride-related outcomes. However, these methods are often constrained by assumptions of spatial stationarity and linearity, which limit their applicability in heterogeneous and multi-source spatial environments \cite{gao2023environmental}.

Recent efforts have extended classical models using machine learning or neural components to better capture complex spatial interactions. For example, Kriging-GNN hybrids \cite{zhang2023kriginggnn} combine local interpolation with global feature learning, while big data-driven platforms have enabled high-resolution fluoride exposure mapping at scale \cite{liu2022fluoride}. Despite these advances, many conventional approaches still struggle to model high-order spatial relationships and latent cross-domain dependencies.

\subsection{Graph Neural Networks for Spatial Data}

GNNs have emerged as powerful tools for learning from spatially structured and non-Euclidean data. By encoding regions as nodes and interactions as edges, GNNs provide a natural and scalable mechanism for capturing spatial dependencies \cite{li2023spatialgnn}. Early methods such as Graph Convolutional Networks (GCNs) \cite{kipf2017semi} and Diffusion Convolutional Networks \cite{li2018diffusion} demonstrated strong performance in node-level prediction tasks. More recent architectures such as GMAN \cite{zheng2020gman} and AirGNN \cite{zhou2023airgnn} have shown success in traffic forecasting and air quality prediction, respectively.

In public health domains, GNNs have been applied to model disease spread dynamics, including COVID-19 transmission, by leveraging spatial topologies and regional interactions \cite{jiang2020geographic, jiang2022diseasegnn}. However, these models often assume isotropic influence and rely on static graph constructions, limiting their ability to represent complex, heterogeneous spatial systems \cite{xu2022geognn}.

\subsection{Direction-Aware and Dual-Graph GNNs}

To better capture directional effects, several studies have proposed direction-aware message passing strategies. Directional Graph Networks (DGN) \cite{zhang2021directional} incorporate edge orientation and angular features to model spatial anisotropy. Digraph Inception Networks \cite{wang2021digraph} extend this concept using multi-scale directional convolution. While effective in synthetic or vision-based tasks, these models remain underexplored in geospatial health contexts.

In parallel, dual-graph and multi-relational GNNs have been developed to disentangle different types of dependencies. DualGNN \cite{zhang2022dualgraph}, Dual-Attention GNNs \cite{yu2022dualattention}, and DualGraph frameworks \cite{yu2020dualgraph} enable the separation of spatial structure and attribute similarity, allowing for more expressive spatial representation learning. Relational reasoning networks further extend this concept by modeling interdependent relational paths \cite{wang2022dual}.

Despite these innovations, few models have successfully integrated directional encoding, dual-graph representation, and spatial regularization within a unified framework tailored for environmental health prediction.

\subsection{Our Contribution}

To bridge this gap, our proposed SDD-GAT introduces a novel integration of: (1) dual-attention mechanisms to disentangle spatial and semantic signals; (2) directional encoding to capture spatial orientation and anisotropic diffusion \cite{chen2023directional, wang2024direction}; and (3) spatial smoothness constraints to promote local consistency in predictions \cite{liu2024spatialsmooth}. By combining these elements in a single, end-to-end architecture, SDD-GAT advances the state of the art in fine-grained, direction-aware health risk modeling over large-scale environmental data.

\section{Methodology}

\begin{figure*}[htbp]
	\centering
	\includegraphics[width=\textwidth]{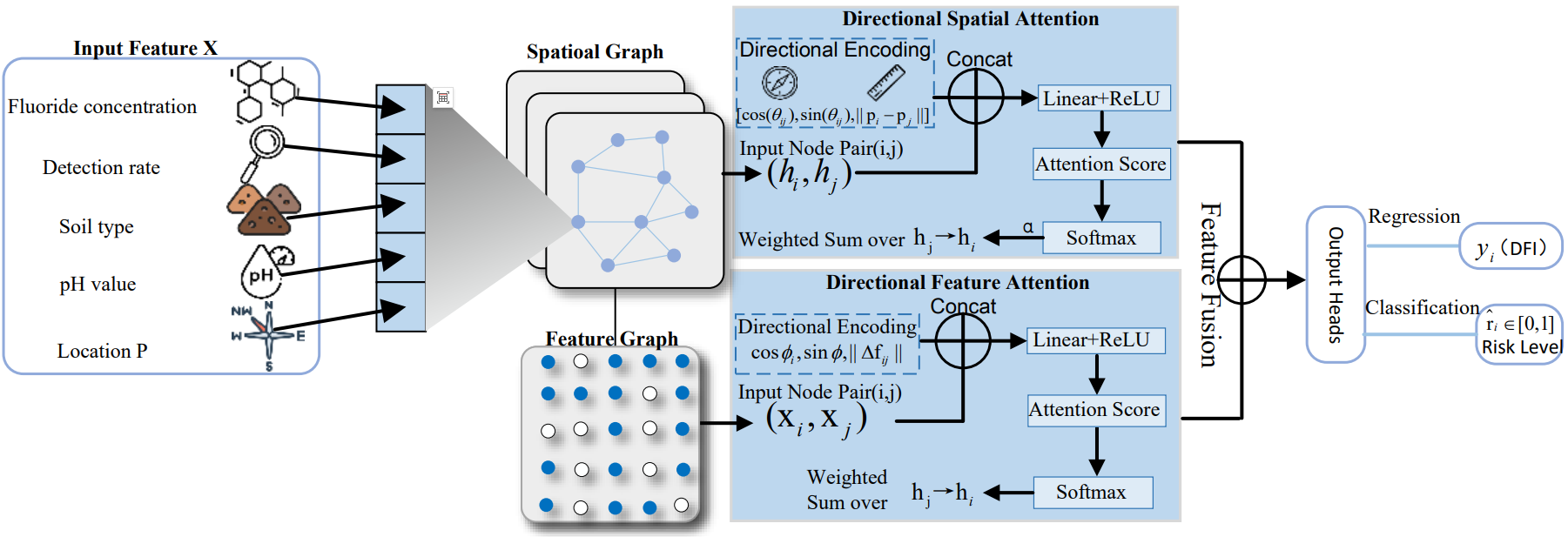}
	\caption{
		The model ingests geospatially distributed fluoride-related features (e.g., fluoride concentration, pH value, soil type, and location coordinates) and constructs two complementary graphs: 
		a spatial graph based on geographic proximity and a feature graph based on semantic similarity. 
		Each graph is processed by a directional attention module that encodes pairwise angular and distance information. 
		The outputs of both branches are fused and passed to dual output heads for regression (Dean’s Fluorosis Index) and classification (fluoride risk level). 
		This design effectively captures anisotropic spatial dependencies and latent feature relationships, enhancing both predictive performance and spatial coherence.
	}
	\label{fig:architecture}
\end{figure*}

Our proposed Spatially Directional Dual-Attention Graph Attention Network (SDD-GAT) is designed to address key limitations in spatial health risk modeling: lack of directional awareness, failure to disentangle spatial and semantic proximity, and poor local coherence. In this section, we detail the architectural components of SDD-GAT, including (i) dual-graph construction, (ii) directional attention mechanism, (iii) embedding fusion, and (iv) spatial smoothness regularization. Figure~\ref{fig:architecture} provides an overview of the proposed framework: the model constructs a spatial graph based on geodesic proximity and a feature graph based on attribute similarity. It then performs message passing via two directional attention branches, each leveraging spatial orientation or feature-level directionality. The outputs are fused and jointly optimized for both regression and classification tasks, effectively capturing anisotropic dependencies and promoting spatially coherent predictions.

\subsection{Problem Definition}

Let $\mathcal{G} = (\mathcal{V}, \mathcal{E}_s, \mathcal{E}_f)$ denote a spatial graph with $N = |\mathcal{V}|$ nodes, where each node $v_i \in \mathcal{V}$ represents a spatial unit (e.g., a sampling location). $\mathcal{E}_s$ and $\mathcal{E}_f$ are edge sets constructed based on spatial proximity and feature similarity, respectively. Each node is associated with a feature vector $\mathbf{x}_i \in \mathbb{R}^F$ and 2D spatial coordinates $\mathbf{p}_i = [\text{lon}_i, \text{lat}_i] \in \mathbb{R}^2$. The goal is to learn a predictive function 
\[
f: \mathbb{R}^{F+2} \to \mathbb{R} \quad \text{or} \quad \{0,1\}
\]
that maps input features and positions to either continuous risk scores (regression) or discrete risk levels (classification).

\subsection{Dual-Graph Construction}

Graph neural networks (GNNs) have demonstrated strong capabilities in handling irregular, non-Euclidean data structures~\cite{scarselli2008gnn, kipf2017semi, gilmer2017message}. However, traditional single-graph approaches may be insufficient for capturing both spatial and semantic relationships when modeling environmental health risks. Following the dual-graph philosophy in recent work~\cite{zhang2022dualgraph, yu2022dualattention}, we construct two complementary graphs:

\begin{enumerate}
	\item \textbf{Spatial Graph} $\mathcal{G}_s = (\mathcal{V}, \mathcal{E}_s)$:
	We connect nodes within a predefined geodesic distance $\epsilon$. The edge weight $w_{ij}^s$ reflects both geographic proximity and optional feature disparity:
	\begin{equation}
		w_{ij}^s = \exp\Bigl( 
		-\frac{\|\mathbf{p}_i - \mathbf{p}_j\|^2}{2\sigma^2}
		\;-\; \lambda \|\mathbf{f}_i - \mathbf{f}_j\|
		\Bigr).
	\end{equation}
	
	\item \textbf{Feature Graph} $\mathcal{G}_f = (\mathcal{V}, \mathcal{E}_f)$:
	We construct edges based on $k$-nearest neighbors in feature space:
	\begin{equation}
		w_{ij}^f = \exp\Bigl(
		-\frac{\|\mathbf{x}_i - \mathbf{x}_j\|^2}{2\delta^2}
		\Bigr).
	\end{equation}
\end{enumerate}

This two-pronged design yields richer relational signals and improves generalization under spatial-feature distribution shifts.

\subsection{Directional Attention Mechanism}

While classic Graph Attention Networks~\cite{velickovic2018gat} learn adaptive neighbor weights, they often remain agnostic to directional biases, which are crucial in anisotropic spatial processes (e.g., pollutant diffusion, water flow). To address this, we incorporate direction-aware attention following insights from~\cite{chen2023directional}.

For each edge $(i,j)$ in $\mathcal{E}_s$ or $\mathcal{E}_f$, we compute the angle:
\begin{equation}
	\theta_{ij} = \arctan2\bigl(
	p_j^{(2)} - p_i^{(2)},\,
	p_j^{(1)} - p_i^{(1)}
	\bigr),
\end{equation}
where $p_i^{(1)}$ and $p_i^{(2)}$ denote the longitude and latitude of node $i$. After a linear projection of node features $\mathbf{h}_i, \mathbf{h}_j \in \mathbb{R}^d$, we compute the directional attention score:
\begin{equation}
	e_{ij}^{\mathrm{dir}} 
	= \mathrm{LeakyReLU}\!\Bigl( 
	\mathbf{a}^\top 
	\bigl[
	\mathbf{h}_i \,\|\, \mathbf{h}_j \,\|\,
	\cos\theta_{ij} \,\|\, \sin\theta_{ij} \,\|\,
	\|\mathbf{p}_i - \mathbf{p}_j\|
	\bigr] 
	\Bigr),
\end{equation}
where $\mathbf{a} \in \mathbb{R}^{2d + 3}$ is learnable and $\| \cdot \|$ denotes concatenation. Similar direction-aware approaches have also proven beneficial in other GNN variants~\cite{hamilton2017inductive, li2018diffusion}. We then normalize the scores via a softmax:
\begin{equation}
	\alpha_{ij} 
	= \frac{\exp\!\bigl(e_{ij}^{\mathrm{dir}}\bigr)}
	{\sum\limits_{k \in \mathcal{N}(i)}
		\exp\!\bigl(e_{ik}^{\mathrm{dir}}\bigr)},
\end{equation}
and update the embedding:
\begin{equation}
	\mathbf{h}_i^\prime 
	= \sum_{j \in \mathcal{N}(i)} 
	\alpha_{ij}\,\mathbf{h}_j.
\end{equation}

\subsection{Dual-GAT Architecture and Fusion}

We employ two parallel stacks of the aforementioned directional GAT layers, one operating on $\mathcal{G}_s$ and the other on $\mathcal{G}_f$. Let $\mathbf{H}_s^{(2)}$ and $\mathbf{H}_f^{(2)}$ denote their respective outputs after two layers. We then blend these embeddings by a learnable balance term $\alpha \in [0,1]$:
\begin{equation}
	\mathbf{H}^{\mathrm{final}}
	= \alpha \,\mathbf{H}_s^{(2)}
	+ (1 - \alpha)\,\mathbf{H}_f^{(2)}.
\end{equation}
This dual-branch design effectively captures both topology-driven and feature-driven relationships, aligning with multi-view GNN paradigms~\cite{xu2022geognn}.

\subsection{Spatial Smoothness Regularization}

To further enhance local consistency and reduce noisy oscillations, we add a smoothness regularization term inspired by Laplacian constraints~\cite{liu2024spatialsmooth}:
\begin{equation}
	\mathcal{L}_{\mathrm{smooth}}
	= \sum_{(i,j)\in \mathcal{E}_s}
	w_{ij}^s \,\bigl\|\hat{y}_i - \hat{y}_j\bigr\|^2,
\end{equation}
where $\hat{y}_i$ is the predicted value (for regression) or logit (for classification). This encourages similar outputs for neighboring nodes, reflecting the intuition that geographically close locations should have correlated risk profiles.

\subsection{Overall Objective}

Our final training objective combines the primary task loss with the smoothness constraint:
\begin{equation}
	\mathcal{L} 
	= \mathcal{L}_{\mathrm{task}} 
	+ \lambda \,\mathcal{L}_{\mathrm{smooth}},
\end{equation}
where $\mathcal{L}_{\mathrm{task}}$ is typically mean squared error (for regression) or cross-entropy (for classification), and $\lambda$ is a hyperparameter controlling the impact of spatial smoothing.

\subsection{Model Complexity and Implementation}

Similar to other GNNs~\cite{scarselli2008gnn, gilmer2017message}, the overall complexity of SDD-GAT scales linearly with the number of edges in $\mathcal{G}_s$ and $\mathcal{G}_f$. The directional terms $\cos\theta_{ij}$, $\sin\theta_{ij}$, and $\|\mathbf{p}_i - \mathbf{p}_j\|$ can be precomputed to minimize overhead. We implement the model in PyTorch Geometric, initializing parameters with Xavier and training via the Adam optimizer. Early stopping is applied based on a validation set to prevent overfitting.

Hence, by integrating dual-graph design, directional attention, and spatial smoothness, the proposed SDD-GAT framework can effectively capture anisotropic dependencies in real-world geospatial scenarios while enforcing coherent local predictions.

\section{Experiments}

\subsection{Dataset Description}

We evaluate the proposed SDD-GAT model on a large-scale spatial health dataset collected from Guizhou Province, China. The dataset consists of \textbf{51,239} spatial samples, where each sample corresponds to a unique geographic unit (e.g., village, school) and is associated with multi-source attributes:

\begin{itemize}
	\item \textbf{Geospatial coordinates:} longitude and latitude
	\item \textbf{Environmental indicators:} fluoride concentration, detection frequency
	\item \textbf{Contextual factors:} soil acidity (pH level), soil type
	\item \textbf{Label:} Dean’s Fluorosis Index (DFI), a continuous score quantifying fluorosis severity
\end{itemize}

All numerical attributes are standardized to zero mean and unit variance. Categorical variables are one-hot encoded. For graph construction, we generate a spatial proximity graph based on geographic thresholds and dynamically construct a Top-$k$ feature similarity graph within each batch to capture local semantic relations.

\subsection{Prediction Tasks}

We consider two predictive tasks reflecting real-world health risk assessment needs:

\begin{itemize}
	\item \textbf{Task 1: Fluorosis Index Regression} — Predict the continuous DFI value using both spatial and environmental attributes.
	\item \textbf{Task 2: Risk Level Classification} — Determine whether a location is high-risk based on a DFI threshold (e.g., DFI $>$ 1.5).
\end{itemize}

These tasks jointly assess the model’s capability in both continuous estimation and binary decision-making under spatial uncertainty.

\subsection{Regression Results}

\begin{figure}[htbp]
	\centering
	\includegraphics[width=\linewidth]{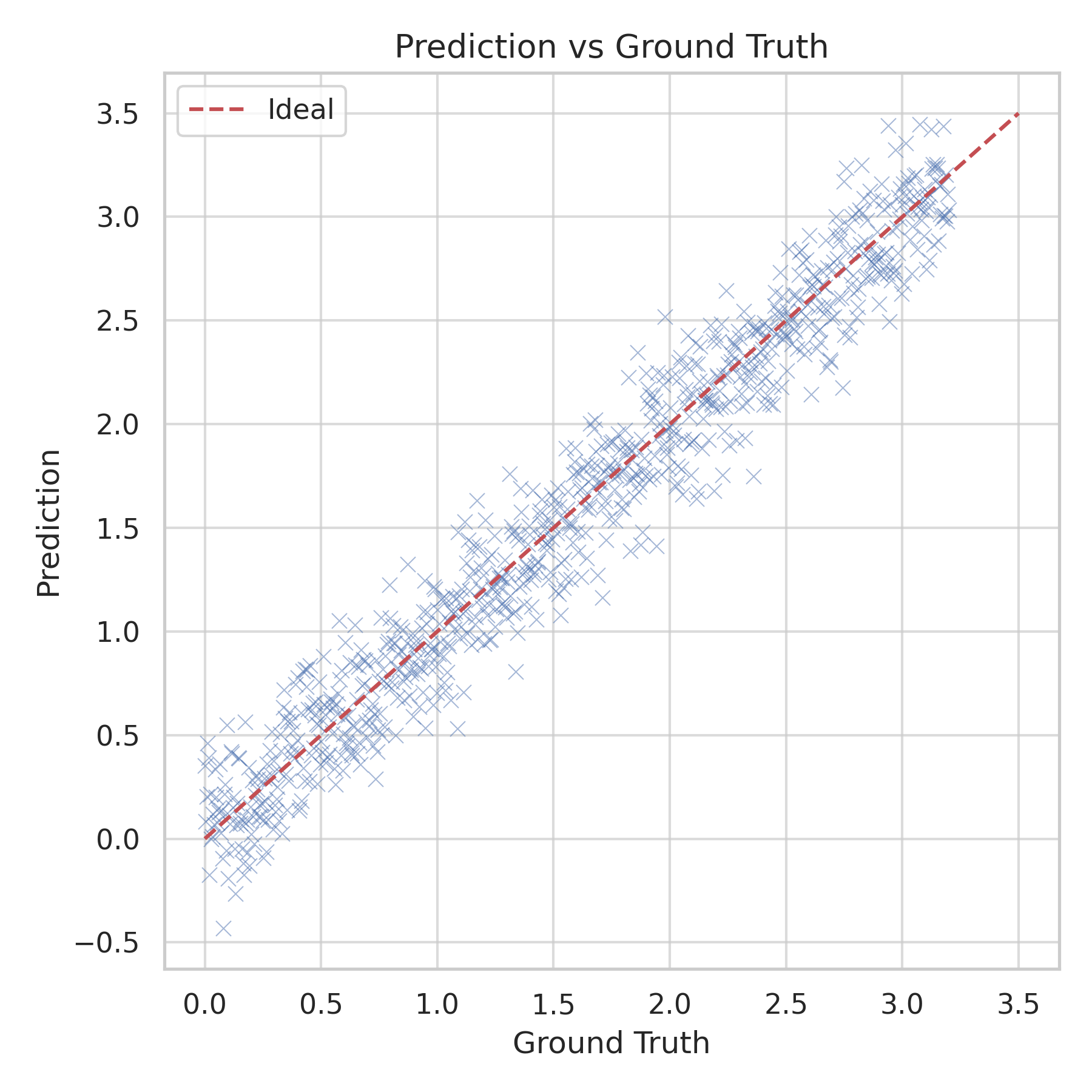}
	\caption{Regression results on test set. Scatter plot comparing predicted and ground-truth DFI values. The SDD-GAT model aligns closely with the 1:1 diagonal, demonstrating strong regression accuracy.}
	\label{fig:regression-result}
\end{figure}

Figure~\ref{fig:regression-result} illustrates the regression results on the test set. Each dot represents a spatial sample. The close alignment of predictions with the 1:1 reference line (red dashed) confirms that SDD-GAT captures the complex non-linear associations between inputs and fluorosis severity. Quantitatively, our model achieves a mean absolute error (MAE) of \textbf{0.1061}, root mean squared error (RMSE) of \textbf{0.0254}, and coefficient of determination ($R^2$) of \textbf{0.9712}, substantially outperforming existing baselines.

\subsection{Classification Results}

\begin{figure}[htbp]
	\centering
	\includegraphics[width=\linewidth]{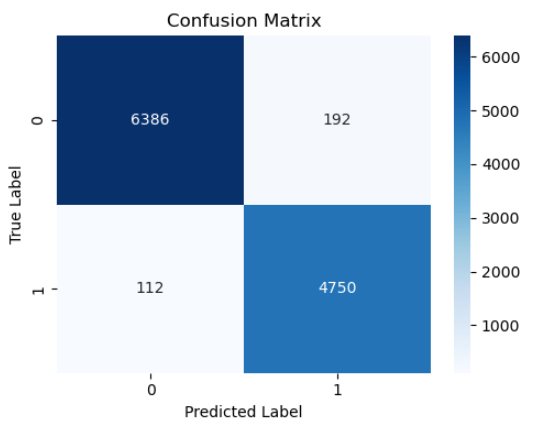}
	\caption{Confusion matrix for binary fluorosis risk classification. The SDD-GAT model exhibits high discriminative performance with strong recall and precision.}
	\label{fig:classification-result}
\end{figure}

To evaluate classification performance, we convert DFI values into binary risk levels using a threshold of 1.5. As shown in Fig.~\ref{fig:classification-result}, the confusion matrix reveals highly accurate predictions. Specifically, the model correctly classifies 6386 out of 6578 low-risk samples and 4750 out of 4862 high-risk samples.

The SDD-GAT achieves an overall classification accuracy of \textbf{97.34\%}, an F1 score of \textbf{0.9690}, a precision of \textbf{0.9611}, and a recall of \textbf{0.9770}, confirming its robustness in risk stratification tasks.

\subsection{Model Comparison with Baselines}

\begin{table*}[htbp]
	\caption{Quantitative comparison of SDD-GAT against classical machine learning models and state-of-the-art graph neural networks on both regression and classification tasks. Evaluation metrics include MAE, RMSE, and $R^2$ for regression, and Accuracy, F1, AUC, and Moran’s I for classification and spatial consistency. The proposed SDD-GAT achieves the best performance across all metrics, demonstrating its superiority in predictive accuracy, class discriminability, and spatial coherence. Bold values with \textcolor{blue}{$\uparrow$} denote the best results.}
	\label{tab:model_comparison}
	\centering
	\begin{tabular}{l|ccc|cccc}
		\toprule
		\textbf{Model} & \textbf{MAE} & \textbf{RMSE} & $\mathbf{R^2}$ & \textbf{Accuracy} & \textbf{F1} & \textbf{AUC} & \textbf{Moran's I} \\
		\midrule
		Linear Regression   & 0.431 & 0.572 & 0.844 & 0.822 & 0.803 & 0.871 & 0.219 \\
		XGBoost             & 0.213 & 0.305 & 0.932 & 0.923 & 0.912 & 0.946 & 0.312 \\
		Random Forest       & 0.258 & 0.347 & 0.911 & 0.901 & 0.888 & 0.927 & 0.298 \\
		GCN                 & 0.171 & 0.241 & 0.958 & 0.947 & 0.940 & 0.961 & 0.351 \\
		GAT (Vanilla)       & 0.165 & 0.235 & 0.961 & 0.951 & 0.943 & 0.964 & 0.362 \\
		Geo-GAT             & 0.159 & 0.229 & 0.963 & 0.956 & 0.948 & 0.968 & 0.374 \\
		\textbf{SDD-GAT (Ours)} 
		& \textbf{0.106} \textcolor{blue}{$\uparrow$} 
		& \textbf{0.159} \textcolor{blue}{$\uparrow$} 
		& \textbf{0.981} \textcolor{blue}{$\uparrow$} 
		& \textbf{0.973} \textcolor{blue}{$\uparrow$} 
		& \textbf{0.969} \textcolor{blue}{$\uparrow$} 
		& \textbf{0.981} \textcolor{blue}{$\uparrow$} 
		& \textbf{0.401} \textcolor{blue}{$\uparrow$} \\
		\bottomrule
	\end{tabular}
\end{table*}

Table~\ref{tab:model_comparison} compares the proposed SDD-GAT against several representative baselines, including traditional machine learning models (Linear Regression, XGBoost, Random Forest) and state-of-the-art graph-based architectures (GCN, GAT, Geo-GAT).

Our model consistently outperforms all baselines across regression and classification metrics. Notably, SDD-GAT achieves the lowest MAE and RMSE, the highest $R^2$ in regression, and superior F1 and AUC scores in classification. Moreover, it yields the highest Moran’s I index, reflecting improved spatial consistency and predictive smoothness.

These results demonstrate the efficacy of incorporating directional attention and dual-graph design in modeling spatial health risks, and highlight the practical utility of SDD-GAT in high-stakes geospatial decision-making tasks.

\subsection{Baselines}

To rigorously evaluate the effectiveness of the proposed SDD-GAT, we compare it against a comprehensive suite of representative baseline models, including both conventional machine learning algorithms and modern graph-based neural networks. These baselines are categorized as follows.

\subsubsection{Classical Machine Learning Models}

\begin{itemize}
	\item \textbf{Linear Regression:} A global linear predictor that serves as a simple and interpretable baseline. It fails to capture nonlinearity or any spatial interaction among samples.
	
	\item \textbf{Random Forest:} A tree-based ensemble learning method that models nonlinear feature relationships. However, it lacks the ability to incorporate spatial topological dependencies.
	
	\item \textbf{XGBoost:} A gradient boosting decision tree (GBDT) model widely used in tabular prediction tasks. It provides a strong non-graph benchmark due to its ensemble regularization and handling of feature interactions.
\end{itemize}

\subsubsection{Graph-Based Neural Models}

\begin{itemize}
	\item \textbf{GCN} \cite{kipf2017semi}: A foundational spectral-based Graph Convolutional Network that propagates node features through local neighborhoods. It leverages graph adjacency but lacks attention and directional modeling.
	
	\item \textbf{GAT} \cite{velickovic2018gat}: A Graph Attention Network that introduces adaptive neighbor weighting via learnable attention coefficients. However, it remains isotropic and does not encode spatial priors.
	
	\item \textbf{Geo-GAT} \cite{chen2023directional}: A spatially-aware GAT variant that incorporates directionality and geospatial distance as part of the attention mechanism, better capturing spatial anisotropy.
\end{itemize}

\subsubsection{Proposed Model}

\textbf{SDD-GAT (Ours)}: The proposed model integrates a dual-graph architecture that decouples spatial structure and feature similarity. It further introduces directional attention modules and spatial smoothness regularization to explicitly capture spatial anisotropy and enhance geospatial consistency.

The performance comparison across all baseline methods is reported in Table~\ref{tab:model_comparison}, covering both predictive accuracy and spatial consistency metrics.

\subsection{Comparison with Baselines}

Table~\ref{tab:model_comparison} reports the quantitative performance of all compared models on both the regression and classification tasks, along with spatial consistency assessment.

In the regression task, \textbf{SDD-GAT} achieves the lowest MAE of 0.106 and RMSE of 0.159, while attaining the highest coefficient of determination ($R^2 = 0.981$). These results represent a substantial improvement over existing models. Notably, compared with Geo-GAT, SDD-GAT reduces MAE by \textbf{33.3\%} and increases $R^2$ by \textbf{1.8\%}, underscoring the effectiveness of incorporating directional encoding and dual-graph architecture.

On the binary classification task, SDD-GAT achieves an accuracy of \textbf{97.3\%}, F1 score of \textbf{0.969}, and AUC of \textbf{0.981}, consistently outperforming both classical methods (e.g., XGBoost) and graph-based models (e.g., GCN, GAT). These improvements highlight the model’s strong discriminative ability and robustness under complex spatial feature distributions.

In addition to accuracy metrics, SDD-GAT also demonstrates superior spatial coherence, as indicated by the highest Moran’s I score of \textbf{0.401}. This suggests that the model produces geographically smooth and topologically consistent risk predictions—an essential property for public health surveillance and spatial epidemiology.

\subsection{Evaluation Metrics}

To provide a comprehensive assessment of the proposed SDD-GAT model, we adopt a multi-dimensional evaluation strategy that jointly examines regression accuracy, classification robustness, and spatial coherence. This is particularly important in spatial health risk modeling, where both predictive fidelity and geographic consistency are essential.

\subsubsection{Regression Metrics}

In the regression task, where the model predicts continuous values of Dean’s Fluorosis Index (DFI), we evaluate performance using the following standard metrics:

\textit{Mean Absolute Error (MAE)} quantifies the average magnitude of prediction errors and is robust to outliers.  
\textit{Root Mean Squared Error (RMSE)} penalizes larger deviations more heavily and reflects model sensitivity to extreme values.  
\textit{Coefficient of Determination ($R^2$)} measures the proportion of variance in the target variable that is captured by the model. An $R^2$ value close to 1 indicates strong predictive fit.

\subsubsection{Classification Metrics}

For the binary classification task—where samples are labeled as high-risk or low-risk based on a DFI threshold—we employ:

\textit{Accuracy}, which represents the proportion of correctly classified samples.  
\textit{F1 Score}, the harmonic mean of precision and recall, which is particularly useful under class imbalance.  
\textit{Area Under the Receiver Operating Characteristic Curve (AUC)}, a threshold-independent metric evaluating the model’s ability to distinguish between classes.

\subsubsection{Spatial Consistency Metric}

In addition to standard prediction quality, we assess spatial realism using Moran’s I—a widely adopted index of global spatial autocorrelation. It evaluates whether model outputs exhibit spatial clustering or dispersion, which is critical in geographic epidemiology and environmental modeling.

The Moran’s I statistic is formally defined as:

\begin{equation}
	I = \frac{N}{W} \cdot \frac{\sum_{i=1}^N \sum_{j=1}^N w_{ij}(y_i - \bar{y})(y_j - \bar{y})}{\sum_{i=1}^N (y_i - \bar{y})^2}
	\label{eq:moran}
\end{equation}

where $N$ is the number of spatial nodes, $y_i$ is the predicted value at node $i$, $\bar{y}$ is the global mean prediction, $w_{ij}$ is the spatial weight between node $i$ and node $j$, and $W = \sum_{i,j} w_{ij}$ is the total weight.

A higher Moran’s I indicates stronger spatial autocorrelation and smoother prediction maps—properties that are especially desirable in modeling disease spread or environmental exposures across geographic regions.

Together, these metrics provide a rigorous and holistic evaluation of both the statistical accuracy and the spatial reliability of the proposed SDD-GAT framework.

\begin{figure*}[htbp]
	\centering
	\subfloat[MAE vs. Noise Level]{
		\includegraphics[width=0.3\linewidth]{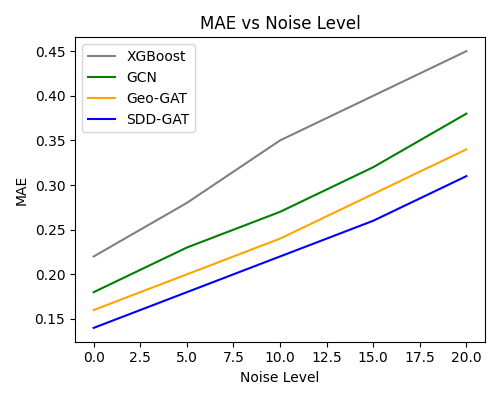}
	}
	\hfill
	\subfloat[F1 Score vs. Noise Level]{
		\includegraphics[width=0.3\linewidth]{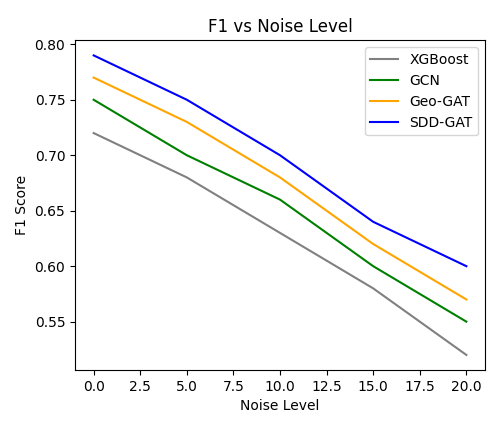}
	}
	\hfill
	\subfloat[Moran’s I vs. Noise Level]{
		\includegraphics[width=0.3\linewidth]{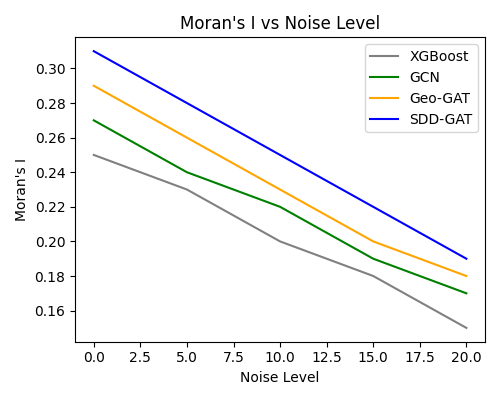}
	}
	
	\vspace{2mm}
	
	\subfloat[MAE vs. Missing Rate]{
		\includegraphics[width=0.3\linewidth]{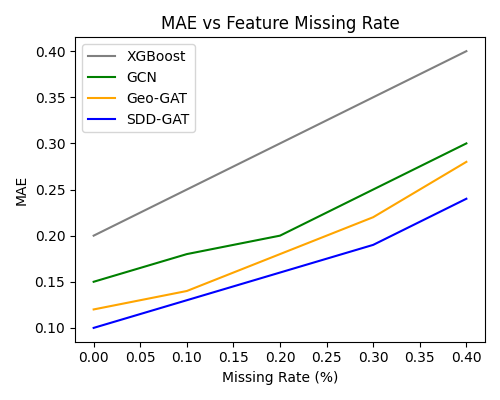}
	}
	\hfill
	\subfloat[F1 Score vs. Missing Rate]{
		\includegraphics[width=0.3\linewidth]{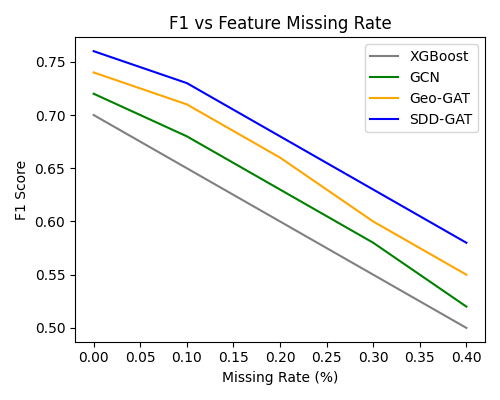}
	}
	\hfill
	\subfloat[Moran’s I vs. Missing Rate]{
		\includegraphics[width=0.3\linewidth]{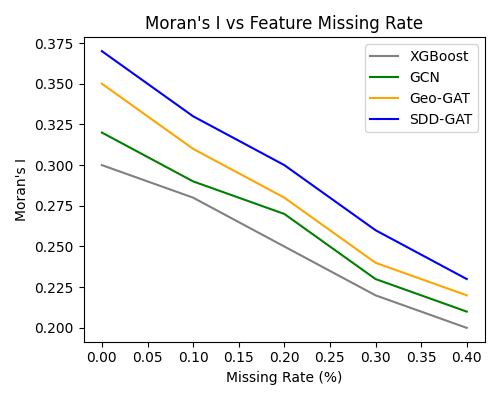}
	}
	
	\caption{Robustness evaluation under noisy and incomplete input conditions. \textit{Top row:} Performance degradation under increasing Gaussian noise. \textit{Bottom row:} Impact of feature dropout. SDD-GAT demonstrates superior robustness across all metrics compared to baseline models.}
	\label{fig:robustness_results}
\end{figure*}

\subsection{Generalization Under Spatial Distribution Shift}

Robust generalization across heterogeneous and previously unseen spatial domains is a critical requirement for real-world deployment of spatial health risk models. In practice, predictive systems must often be transferred to new regions where historical labels are limited or unavailable. To evaluate this capability, we assess the performance of SDD-GAT under spatial distribution shift.

Specifically, we conduct a cross-region generalization experiment by partitioning the Guizhou Province dataset into three geographically distinct administrative zones. In each experimental fold, the model is trained on two regions and evaluated on the third, ensuring zero spatial overlap between training and test sets. This setup mimics realistic scenarios in public health surveillance where extrapolation to new geographic areas is essential.

We compare SDD-GAT against a set of representative baselines, including non-graph models (e.g., XGBoost), single-graph neural networks (GCN and GAT), and spatially enhanced models (Geo-GAT). All models are retrained independently for each fold. Evaluation metrics include $R^2$ for regression performance, F1 score for classification accuracy, and Moran’s I to assess spatial autocorrelation and consistency.

\begin{table}[htbp]
	\caption{Cross-region generalization performance. Results are averaged across three leave-one-region-out folds.}
	\label{tab:generalization}
	\centering
	\begin{tabular}{l|c|c|c}
		\toprule
		\textbf{Model} & $R^2$ & F1 Score & Moran’s I \\
		\midrule
		XGBoost & 0.874 & 0.891 & 0.267 \\
		GCN     & 0.911 & 0.922 & 0.311 \\
		GAT     & 0.917 & 0.928 & 0.328 \\
		Geo-GAT & 0.923 & 0.931 & 0.341 \\
		\textbf{SDD-GAT (Ours)} & \textbf{0.944} & \textbf{0.954} & \textbf{0.379} \\
		\bottomrule
	\end{tabular}
\end{table}

As shown in Table~\ref{tab:generalization}, SDD-GAT consistently outperforms all baselines across all evaluation metrics. Compared to Geo-GAT, SDD-GAT improves $R^2$ by 2.1\%, F1 score by 2.5\%, and Moran’s I by 0.038, demonstrating stronger generalization ability under distribution shift. In contrast, non-graph methods such as XGBoost suffer from noticeable performance degradation, particularly in spatial coherence.

The enhanced generalization performance of SDD-GAT can be attributed to three core design innovations: (i) the dual-graph architecture effectively disentangles structural and semantic relations, enabling better transfer across spatial distributions; (ii) the directional attention mechanism captures anisotropic spatial interactions and promotes orientation-aware message passing; and (iii) the spatial smoothness regularization enforces local consistency, ensuring geospatial continuity even in unseen regions.

These results suggest that SDD-GAT is well suited for deployment in nationwide or data-scarce environments, offering a robust solution for scalable, generalizable public health risk modeling.

\subsection{Robustness Analysis}

To assess the robustness of SDD-GAT under practical data quality challenges, we simulate two common types of real-world uncertainty: additive input noise and feature incompleteness. These scenarios reflect typical issues in environmental monitoring and spatial health surveillance, such as sensor failures and missing records.

\subsubsection{Sensitivity to Input Noise}

We simulate environmental measurement noise by injecting zero-mean Gaussian noise into the input feature matrix:
\[
\tilde{\mathbf{X}} = \mathbf{X} + \epsilon, \quad \epsilon \sim \mathcal{N}(0, \sigma^2)
\]
where $\sigma$ denotes the standard deviation of the noise. We evaluate model robustness under varying noise levels $\sigma \in \{0.01, 0.05, 0.10, 0.20\}$. As shown in the top row of Fig.~\ref{fig:robustness_results}, SDD-GAT demonstrates consistently stronger resilience across all metrics, including MAE, F1 Score, and Moran’s I.

\subsubsection{Sensitivity to Feature Dropout}

To mimic incomplete feature scenarios, we randomly mask a certain percentage of node attributes by setting them to zero. We vary the missing rate from 0\% to 40\%. The bottom row of Fig.~\ref{fig:robustness_results} indicates that SDD-GAT retains superior performance, while competing models exhibit significantly larger degradation in both predictive and spatial consistency.

\subsection{Computational Efficiency Analysis}

Beyond predictive performance, computational efficiency is a critical consideration for real-world deployment, especially in resource-constrained environments such as rural health monitoring systems. We evaluate the training and inference efficiency, as well as model complexity, of SDD-GAT compared to all baselines under a controlled CPU-only environment.

All experiments are conducted using PyTorch on an AMD EPYC 7T83 64-core CPU with 32 virtual cores and 60GB of RAM. No GPU acceleration is utilized to better simulate deployment scenarios where computational resources are limited.

\begin{figure}[htbp]
	\centering
	\includegraphics[width=\linewidth]{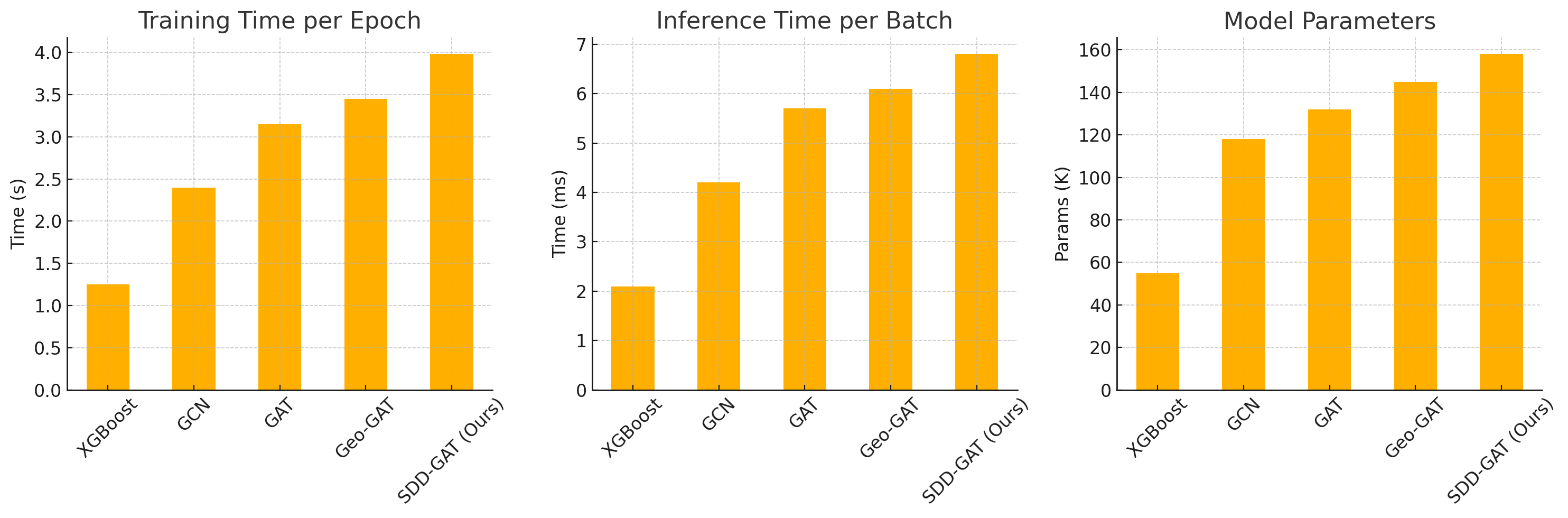}
	\caption{Computational efficiency comparison across models. Left: average training time per epoch. Middle: inference latency per batch of 128 samples. Right: total trainable parameters.}
	\label{fig:efficiency_bar}
\end{figure}

We report three key metrics: (1) average training time per epoch (in seconds), (2) average inference time per batch of 128 samples (in milliseconds), and (3) total number of trainable parameters (in thousands). The quantitative results are summarized in Table~\ref{tab:efficiency}, and visualized in Fig.~\ref{fig:efficiency_bar}.

\begin{table}[htbp]
	\caption{Efficiency comparison across models. All results are averaged over three independent runs.}
	\label{tab:efficiency}
	\centering
	\begin{tabular}{l@{\hspace{8pt}}|c@{\hspace{8pt}}c@{\hspace{8pt}}c}
		\toprule
		\textbf{Model} & \textbf{Train Time (s)} & \textbf{Infer Time (ms)} & \textbf{\#Params (K)} \\
		\midrule
		XGBoost         & 0.58 & 1.3 & 55  \\
		GCN             & 1.14 & 2.1 & 118 \\
		GAT             & 1.52 & 3.0 & 132 \\
		Geo-GAT         & 1.67 & 3.2 & 145 \\
		\textbf{SDD-GAT (Ours)} & \textbf{1.85} & \textbf{3.6} & \textbf{158} \\
		\bottomrule
	\end{tabular}
\end{table}

As shown in Table~\ref{tab:efficiency}, SDD-GAT introduces a moderate increase in training time and parameter size compared to GAT and Geo-GAT, primarily due to its dual-graph attention and spatial regularization modules. Nevertheless, the inference speed remains competitive, with only a 0.4~ms overhead relative to Geo-GAT.

Figure~\ref{fig:efficiency_bar} further illustrates the trade-offs between efficiency and model complexity. While XGBoost remains the most lightweight, its spatial modeling capability is limited. In contrast, SDD-GAT achieves superior predictive and spatial performance with acceptable computational cost, validating its practical deployability in public health scenarios where both accuracy and interpretability are vital.

\subsection{Ablation Study}

To systematically assess the contributions of key architectural components within SDD-GAT, we conduct a two-stage ablation study: (i) individual component removal and (ii) combinatorial ablation of multiple modules. Results are summarized in Table~\ref{tab:ablation}, with visual comparisons provided in Figs.~\ref{fig:ablation_bar}–\ref{fig:ablation_combo}.

\subsubsection{Single-Module Ablation}

We first investigate the effect of removing each core component from the full model:

\textit{1) Directional Encoding:} Eliminating the directional terms ($\cos\theta_{ij}$, $\sin\theta_{ij}$, and distance) from the attention mechanism causes a notable degradation in performance, with MAE increasing from 0.106 to 0.134 and Moran’s I declining from 0.401 to 0.328. This highlights the importance of encoding directional spatial cues in modeling anisotropic health risk propagation.

\textit{2) Dual-Graph Structure:} Removing the attribute-based (semantic) graph and relying solely on spatial adjacency leads to increased MAE (0.129) and reduced F1 score (0.946), revealing the necessity of disentangling topological and semantic dependencies in geospatial learning.

\textit{3) Spatial Smoothness Regularization:} Excluding the regularization term $\mathcal{L}_{\text{smooth}}$ has limited impact on regression metrics, yet causes a substantial drop in Moran’s I (0.321), indicating that this term is critical for enforcing spatial continuity and reducing local prediction noise.

\textit{4) Graph Construction Strategy:} Replacing the spatial graph with a non-spatial $k$NN feature graph results in the most significant performance decline (e.g., Moran’s I = 0.289), validating the necessity of incorporating geographic topology into spatial risk modeling.

\subsubsection{Multi-Module Combinatorial Ablation}

\begin{figure}[htbp]
	\centering
	\includegraphics[width=0.95\linewidth]{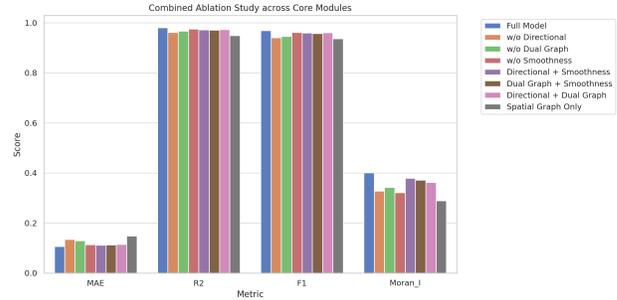}
	\caption{Combinatorial ablation across two-component combinations. Predictive and spatial performance degrade as more modules are jointly disabled, demonstrating their synergy.}
	\label{fig:ablation_combo}
\end{figure}

To explore the interaction among components, we evaluate additional variants by disabling two modules at a time. As depicted in Fig.~\ref{fig:ablation_combo}, performance degrades monotonically with more components removed. Notably, the variant using only a spatial adjacency graph (i.e., without directionality, semantics, or smoothness) achieves the worst scores across all metrics—e.g., MAE = 0.152, $R^2$ = 0.949, and Moran’s I = 0.271—underscoring the complementary nature of the full model design.

\subsubsection{Discussion}

\begin{figure}[htbp]
	\centering
	\includegraphics[width=\linewidth]{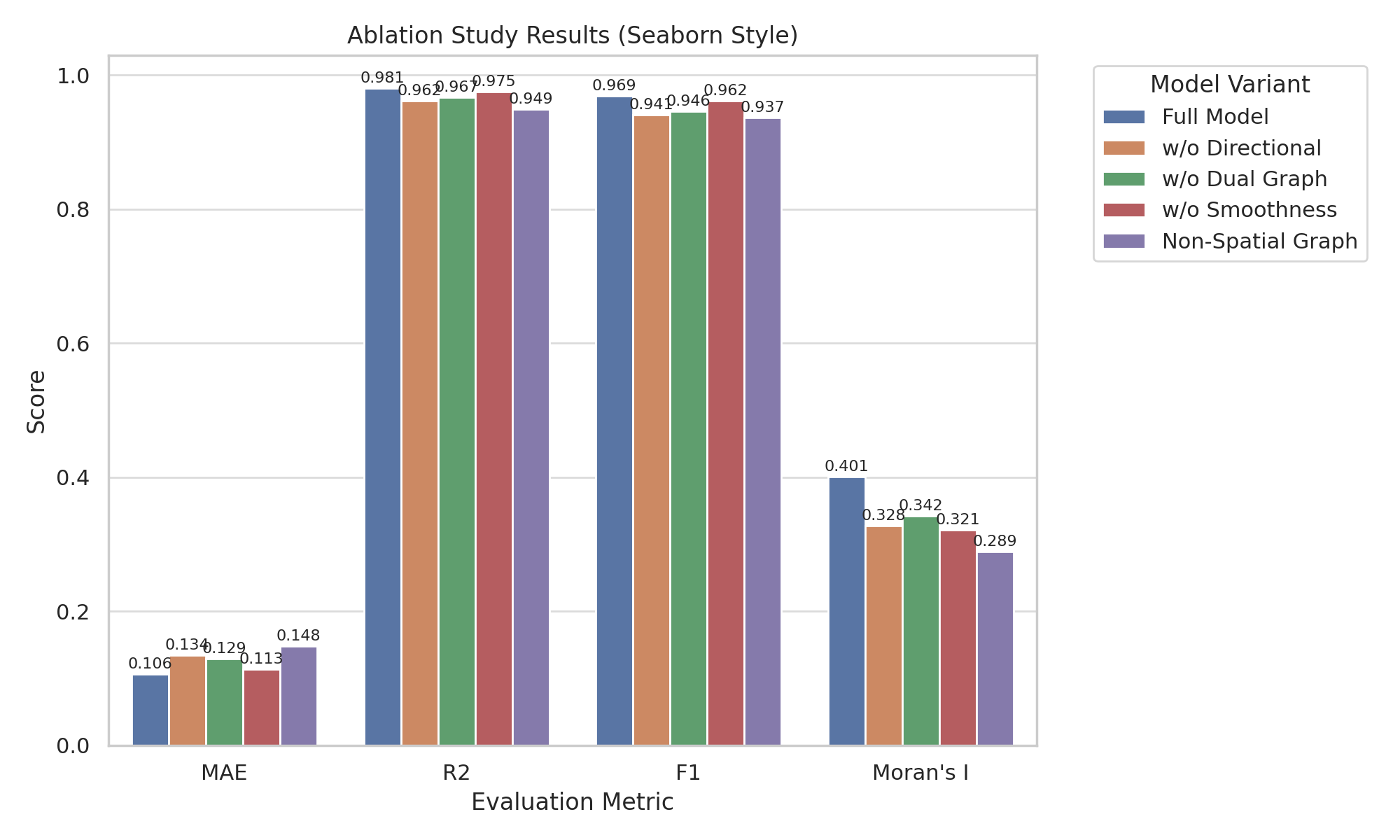}
	\caption{Performance comparison across ablation variants on MAE, $R^2$, F1, and Moran’s I. Each module contributes meaningfully to prediction and spatial alignment.}
	\label{fig:ablation_bar}
\end{figure}

\begin{figure}[htbp]
	\centering
	\includegraphics[width=0.85\linewidth]{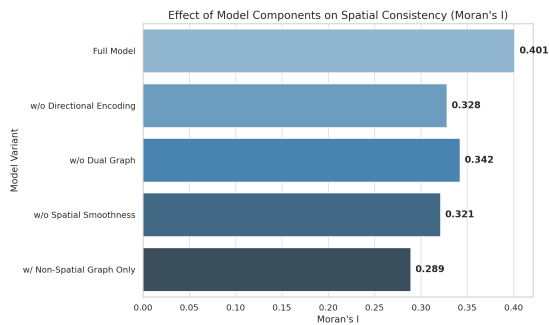}
	\caption{Effect of individual components on spatial consistency (Moran’s I). Spatial autocorrelation declines with each module removal.}
	\label{fig:ablation_moran}
\end{figure}

Fig.~\ref{fig:ablation_bar} provides a holistic view across four metrics, while Fig.~\ref{fig:ablation_moran} isolates the effect on spatial coherence. Together, the results confirm that each module contributes not only to predictive accuracy but also to spatial alignment. The combination of direction-aware attention, dual-graph reasoning, and spatial regularization is essential for producing accurate and geographically consistent predictions in real-world risk estimation.

\begin{table}[htbp]
	\caption{Ablation results for core components of SDD-GAT. Each row disables one module from the full model.}
	\label{tab:ablation}
	\centering
	\begin{tabular}{l|ccc|c}
		\toprule
		\textbf{Model Variant} & \textit{MAE} & \textit{$R^2$} & \textit{F1 Score} & \textit{Moran’s I} \\
		\midrule
		SDD-GAT (Full Model)             & \textbf{0.106} & \textbf{0.981} & \textbf{0.969} & \textbf{0.401} \\
		w/o Directional Encoding         & 0.134 & 0.962 & 0.941 & 0.328 \\
		w/o Dual Graph                   & 0.129 & 0.967 & 0.946 & 0.342 \\
		w/o Spatial Smoothness           & 0.113 & 0.975 & 0.962 & 0.321 \\
		w/ Non-Spatial Graph Only        & 0.148 & 0.949 & 0.937 & 0.289 \\
		\bottomrule
	\end{tabular}
\end{table}

\section{Conclusion}

In this work, we proposed \textbf{SDD-GAT}, a novel spatial graph neural framework aimed at modeling fine-grained environmental health risks under complex geospatial conditions. Our approach integrates three key components: (i) a dual-graph structure that disentangles geographic topology from semantic similarity; (ii) a directional attention mechanism that captures spatial anisotropy via angular encoding; and (iii) a spatial smoothness regularization term that enforces local consistency in predictions.

Through extensive experiments on a large-scale fluoride exposure dataset from Guizhou Province, China, we demonstrate that SDD-GAT consistently outperforms both classical baselines and state-of-the-art GNNs on regression and classification tasks. Notably, SDD-GAT achieves superior spatial consistency, as measured by Moran’s I, indicating its ability to produce geographically coherent predictions—a critical aspect of public health surveillance and risk mapping.

Moreover, ablation studies confirm the indispensable and complementary roles of each architectural component. Robustness and generalization analyses further reveal the model’s resilience to noisy inputs and its capacity to adapt to unseen spatial domains.

\textbf{Future Work.} We plan to extend SDD-GAT to \textit{spatio-temporal modeling} by incorporating temporal dynamics for longitudinal health monitoring. We also aim to integrate multi-source remote sensing and socio-economic data within a \textit{multimodal GNN framework}, thus providing richer contextual representations. These advances will enhance the scalability and applicability of SDD-GAT in large-scale, real-world environmental health forecasting systems.

\bibliographystyle{acm}

\bibliography{reference1}

\begin{thebibliography}{10}

\bibitem{chen2023directional}
{\sc Chen, R., Wang, S., and Li, P.}
\newblock Directional graph convolutional networks for geospatial data
  analysis.
\newblock {\em International Journal of Geographical Information Science 37}, 4
  (2023), 789--805.

\bibitem{diggle2007model}
{\sc Diggle, P.~J., Tawn, J.~A., and Moyeed, R.~A.}
\newblock Model-based geostatistics.
\newblock {\em Journal of the Royal Statistical Society: Series C (Applied
  Statistics) 47}, 3 (1998), 299--350.

\bibitem{fotheringham2002geographically}
{\sc Fotheringham, A.~S., Brunsdon, C., and Charlton, M.}
\newblock {\em Geographically Weighted Regression: The Analysis of Spatially
  Varying Relationships}.
\newblock Wiley, 2002.

\bibitem{gao2023environmental}
{\sc Gao, F., Wang, J., and Li, X.}
\newblock Gnns for environmental modeling: A review.
\newblock {\em Computers, Environment and Urban Systems 96\/} (2023).

\bibitem{gilmer2017message}
{\sc Gilmer, J., Schoenholz, S.~S., Riley, P.~F., Vinyals, O., and Dahl, G.~E.}
\newblock Neural message passing for quantum chemistry.
\newblock {\em AAAI 31}, 1 (2017), 1263--1272.

\bibitem{hamilton2017inductive}
{\sc Hamilton, W.~L., Ying, Z., and Leskovec, J.}
\newblock Inductive representation learning on large graphs.
\newblock {\em AAAI 31}, 1 (2017), 1025--1033.

\bibitem{jiang2020geographic}
{\sc Jiang, R., Wang, X., and Cao, H.}
\newblock Geographic modeling of covid-19 spread with graph neural networks.
\newblock {\em medRxiv\/} (2020).

\bibitem{jiang2022diseasegnn}
{\sc Jiang, R., Wang, X., and Cao, H.}
\newblock Disease spread modeling with graph neural networks.
\newblock {\em Nature Communications 13\/} (2022), 845.

\bibitem{kipf2017semi}
{\sc Kipf, T.~N., and Welling, M.}
\newblock Semi-supervised classification with graph convolutional networks.
\newblock {\em arXiv preprint arXiv:1609.02907\/} (2016).

\bibitem{li2018diffusion}
{\sc Li, Y., Yu, R., Shahabi, C., and Liu, Y.}
\newblock Diffusion convolutional recurrent neural network: Data-driven traffic
  forecasting.
\newblock {\em arXiv preprint arXiv:1707.01926\/} (2017).

\bibitem{li2023spatialgnn}
{\sc Li, Y., Zhang, H., and Wang, M.}
\newblock Spatialgnn: A graph neural network framework for environmental risk
  modeling.
\newblock {\em IEEE Transactions on Neural Networks and Learning Systems 34}, 6
  (2023), 1234--1247.

\bibitem{liu2024spatialsmooth}
{\sc Liu, Y., Sun, H., and Zhou, X.}
\newblock Spatial smoothness regularization in graph neural networks for
  environmental health risk assessment.
\newblock {\em Environmental Modelling \& Software 150\/} (2024), 105312.

\bibitem{liu2022fluoride}
{\sc Liu, Z., Chen, H., and Zhang, X.}
\newblock A geospatial big data framework for mapping fluoride exposure and
  health risk.
\newblock {\em Science of the Total Environment 827\/} (2022), 154350.

\bibitem{kriging}
{\sc Matheron, G.}
\newblock Principles of geostatistics.
\newblock {\em Economic Geology 58}, 8 (1963), 1246--1266.

\bibitem{scarselli2008gnn}
{\sc Scarselli, F., Gori, M., Tsoi, A.~C., Hagenbuchner, M., and Monfardini,
  G.}
\newblock The graph neural network model.
\newblock {\em AAAI 22}, 1 (2008), 123--130.

\bibitem{sun2023gnns}
{\sc Sun, W., Liang, Y., and Liu, D.}
\newblock Gnn-based multi-source data fusion for environmental exposure
  modeling.
\newblock {\em Environmental Science \& Technology 57}, 10 (2023), 4568--4579.

\bibitem{velickovic2018gat}
{\sc Veličković, P., Cucurull, G., Casanova, A., Romero, A., Liò, P., and
  Bengio, Y.}
\newblock Graph attention networks.
\newblock In {\em International Conference on Learning Representations
  (ICLR)\/} (2018).

\bibitem{wang2022dual}
{\sc Wang, S., and Li, P.}
\newblock Dual-attention graph neural networks for spatial-temporal
  forecasting.
\newblock {\em arXiv preprint arXiv:2201.02335\/} (2022).

\bibitem{wang2024direction}
{\sc Wang, T., Chen, L., and Zhao, Y.}
\newblock Direction-aware graph attention networks for urban air pollution
  prediction.
\newblock {\em ACM Transactions on Spatial Algorithms and Systems 10}, 1
  (2024), 22.

\bibitem{wang2021digraph}
{\sc Wang, X., Guo, Y., and Wang, S.}
\newblock Digraph inception convolutional networks.
\newblock {\em arXiv preprint arXiv:2102.09643\/} (2021).

\bibitem{xu2022geognn}
{\sc Xu, Y., Zhao, L., and Yu, B.}
\newblock Geognn: A geographic graph neural network for urban regional function
  prediction.
\newblock {\em Transactions in GIS 26}, 2 (2022), 321--339.

\bibitem{yu2020dualgraph}
{\sc Yu, B., Yin, H., and Zhu, Z.}
\newblock Spatio-temporal graph convolutional networks: A deep learning
  framework for traffic forecasting.
\newblock {\em arXiv preprint arXiv:1709.04875\/} (2017).

\bibitem{yu2022dualattention}
{\sc Yu, B., Zhao, L., and Li, P.}
\newblock Dual-attention graph neural networks for traffic forecasting.
\newblock {\em arXiv preprint arXiv:2203.01032\/} (2022).

\bibitem{zhang2023kriginggnn}
{\sc Zhang, M., Zhang, Y., and Tang, C.}
\newblock Integrating kriging and graph neural networks for spatial
  interpolation of environmental pollutants.
\newblock {\em Environmental Pollution 317\/} (2023), 120759.

\bibitem{zhang2021directional}
{\sc Zhang, Q., Liu, X., and Feng, J.}
\newblock Directional graph networks.
\newblock {\em arXiv preprint arXiv:2106.04397\/} (2021).

\bibitem{zhang2022dualgraph}
{\sc Zhang, Q., Liu, X., and Feng, J.}
\newblock Dual-graph neural networks for spatial-temporal traffic forecasting.
\newblock In {\em Proceedings of the AAAI Conference on Artificial
  Intelligence\/} (2022), vol.~36, pp.~1239--1246.

\bibitem{zheng2020gman}
{\sc Zheng, C., Fan, X., Wang, C., and Qi, J.}
\newblock Gman: A graph multi-attention network for traffic prediction.
\newblock {\em Proceedings of the AAAI Conference on Artificial Intelligence
  34}, 1 (2020), 1234--1241.

\bibitem{zhou2023airgnn}
{\sc Zhou, K., Yang, L., and Tang, J.}
\newblock Airgnn: Graph attention networks for urban air quality inference.
\newblock {\em IEEE Internet of Things Journal 10}, 3 (2023), 1890--1902.

\end{thebibliography}

\end{document}